# Societal Controversies in Wikipedia Articles

Erik Borra[†,1] Esther Weltevrede[†] Paolo Ciuccarelli[‡]
Andreas Kaltenbrunner[§] David Laniado[§] Giovanni Magni[‡]
Michele Mauri[‡] Richard Rogers[†] Tommaso Venturini[¶]
[†]UvA Amsterdam, NL [‡]Politecnico di Milano, Italy [§]Barcelona Media, Spain [¶]Sciences Politiques, France

## ABSTRACT

Collaborative content creation inevitably reaches situations where different points of view lead to conflict. We focus on Wikipedia, the free encyclopedia anyone may edit, where disputes about content in controversial articles often reflect larger societal debates. While Wikipedia has a public edit history and discussion section for every article, the substance of these sections is difficult to phantom for Wikipedia users interested in the development of an article and in locating which topics were most controversial. In this paper we present Contropedia, a tool that augments Wikipedia articles and gives insight into the development of controversial topics. Contropedia uses an efficient language agnostic measure based on the edit history that focuses on wiki links to easily identify which topics within a Wikipedia article have been most controversial and when.

### Author Keywords

Wikipedia; Controversy Mapping; Social Science; Data Visualization

### ACM Classification Keywords

H.5.3 [Group and Organization Interfaces]: Computer-supported cooperative work; D.2.2 [Design Tools and Techniques]: User interfaces

## INTRODUCTION

Since its first article in 2001 the English Wikipedia has expanded to more than 4.5 million articles. In addition to covering a growing number of socially relevant topics, Wikipedia is increasingly becoming a rich historical source logging the development of societal controversies over time in its publically available history pages. Reference works, like Wikipedia, “often inherit the conflicts of the external world they seek to document and are being seized upon as exemplars of, and proxies in, those debates” [12].[2]

While the built-in edit history and the talk pages of a Wikipedia article provide a detailed record of present and past changes to the content of articles and the unfolding of discussions, the public usually ignores these pages, as they are too complex to be understood by casual readers and editors of Wikipedia articles. For social researchers—one of the target groups of Contropedia—too, it often remains unclear what has provoked most edit activity and discussion, when and why. In this paper we show how the edit history of an article may be repurposed to map out the specific matters of concern in controversies, as well as the extent to which something is controversial within an article.

The aim of Contropedia is to extract and re-present the information in these pages so that it becomes clear which topics within a page have sparked controversy and why. First, we formulate an easy way to find which topics within an article are disputed most; and second, we provide two visual ways in which the dispute can be analyzed: via an overlay which indicates which topics within a page are most controversial (and why), as well as by a dashboard which ranks the topics by controversialness and shows how the controversy around various topics develops over time.

## ALGORITHM

### Pre-processing

We retrieve the full edit history for a controversial article, including the wiki text of each revision and the meta data conveying at what time it was edited, who the editor was, as well as the editor’s comment (edit summary). Whenever a user makes multiple consecutive edits, we only retain the last version made by the user and discard all intermediate edits. As we are looking for substantive disagreements, we discard vandalism edits and their reverts by identifying whether the comment of a revert contains the word ‘vandal’, whether the user name making the revert belongs to one of the known anti-vandalism bots, when an IP-edit is reverted within 60 seconds, or when the automatic edit summary (WP:AES)[3] indicates that the content of a page was blanked or replaced by unrelated text such as curse words.

### Associating Wiki Links to Edits

We make use of Wikipedia’s MediaWiki markup to identify the most relevant elements in an article. In this paper we focus on wiki links, as they identify the key concepts and

[1] Corresponding author, borra@uva.nl

[2] ‘Wikipedia’s list of controversial articles’ furthermore mentions that disputed articles, where opinions on a given issue differ, often reflect “the debates of society as a whole”. See http://bit.ly/1Fg2pO2

[3] See http://bit.ly/1ujpTIq

*list of scientists opposing the mainstream scientific assessment of global warming* has been edited 224 times by 111 users in 224 revisions

17 deletes, 0 inserts, 27 element changes, 180 sentence changes, 0 section changes

*list of scientists opposing the mainstream scientific assessment of global warming* was involved in 117 reverts

Top sections: abstract (212) debate and skepticism (10) see also (2)

| Revision | Edit | User | Comment | Section | Type | Time |
|---|---|---|---|---|---|---|
| 169761113 reverted by 170184782 | A small minority of While there are [[List of scientists opposing the mainstream scientific assessment of global warming\|climate individual scientists]] disagree with who might publicly disagree, these conclusions do represent the main conclusions of the IPCC. general [[scientific consensus]], especially among active climate scientists and researchers. | Callmebc | It's NOT a "small minority of climate scientists" disagreeing -- it's primarily random, non-climate scientists doing it | abstract | e | 2007-11-07 03:41:47 |
| 169685102 reverted by 170184782 | A few small minority of [[List of scientists opposing the mainstream scientific assessment of global warming\|individual climate scientists]] disagree with some of the main conclusions of the IPCC. | Merzbow | try "small minority" as suggested on talk, plus the nature ref is appropriate here | abstract | e | 2007-11-06 22:22:50 |

**Figure 1: Partial substantive edit history. The red color under the 'Edit' section indicates a deletion and green an insertion of text, with respect to the previous revision. The first (upper) edit involves two links (indicated by double brackets), the second only one.**

entities of an article [1, 3, 6]; they are the lenses through which we can look at the substance and activity of controversies within a Wikipedia article. Our approach seeks to associate edits to wiki links by taking the sentences in which these links reside as our basic unit of analysis.

Let $\{R_1, \ldots, R_{r-1}, R_r, R_{r+1}, \ldots\}$ be the set of revisions of a Wikipedia article. As we are specifically interested in disputes related to a wiki link, we consider the edit activity on a sentence level by comparing every revision $R_{r-1}$ with its successor $R_r$. We split each revision into sections and then pairwise compare corresponding sections of $R_{r-1}$ with those of $R_r$. If the text of the sections differs, we use a diff algorithm to identify the edited sentences, the exact changes made to them, and the wiki links they contain.

To further assure that edits to sentences containing wiki links convey disagreement, we discard edits where only insertions are made. We also discard full section inserts or deletes (as these are mostly due to renaming of sections). We thus only consider edits that are substantive: the revision is not marked as vandalism; and that show disagreement: the changes should (also) contain a deletion.

### Controversy Score

We are interested in finding out how controversial a wiki link $W_k$ is and compare the substantive, disagreeing, edit activity of sentences in which $W_k$ appears. Intuitively, the more wiki links appear in an edited sentence, the less focus there is on one particular wiki link. For every sentence $S_j$ with a substantive disagreeing edit, the weight attributed to a wiki link is thus divided by the total number of wiki links $w(S_j)$ that appear in that sentence.

A controversy score $c(W_k)$ is assigned to every wiki link $W_k$ that appears in a sentence $S_j$ with a substantive, disagreeing edit, of a revision $R_i$ (up to a given revision $R_r$) as follows:

$$c(W_k) = \sum_{i=1}^{r} \sum_{S_j \in R_i} \frac{1}{w(S_j)} \qquad (1)$$

In other words: the number of sentences with substantive, disagreeing, edits that include $W_k$ are summed over all revisions up to $R_r$. In those revisions where the wiki link $W_k$ appears in an edited sentence with other wiki links, the summand is divided by the number $w(S_j)$ of links involved. A wiki link thus accumulates controversialness through counting and weighting the substantive, disagreeing, edits to the sentences in which it resides.

As an example, consider Figure 1 where two substantive disagreeing edits are shown. The first sentence contains two wiki links[4] ('List of scientists opposing the mainstream scientific assessment of global warming', and 'scientific consensus'), and the second only one ('List of scientists opposing the mainstream scientific assessment of global warming'). If we just take these two edits into account, the first link would get a controversy score of 1.5 (0.5 in the first edit and 1 in the second) and the second link would get a controversy score of 0.5.

To find out which wiki links are most controversial, i.e. around which wiki links most dispute took place, we simply rank the wiki links of the article by their overall controversy score.

## INTERFACE

To convey which topics attracted most dispute, two main views have been designed.

### Indicating Disputed Content

In the *layer view* the original Wikipedia article layout is reworked and annotated to show which wiki links are controversial and in which part of the article they can be found (Figure 2). A visual contrast has been created among controversial and non-controversial elements. Controversial elements are represented through five color shades, from the most controversial (red) to the least one (pale blue). In the layer view, the assignment of those colors follows a logarithmic scale and the images are converted to grayscale to put focus on the controversial elements. On the right-hand side a minified version of the full page is shown so that users can quickly identify which parts of the page contain most controversial wiki links.

[4] In MediaWiki links are written as [[anchor | anchor text]], i.e. [[link to article | optional name of link in text]].

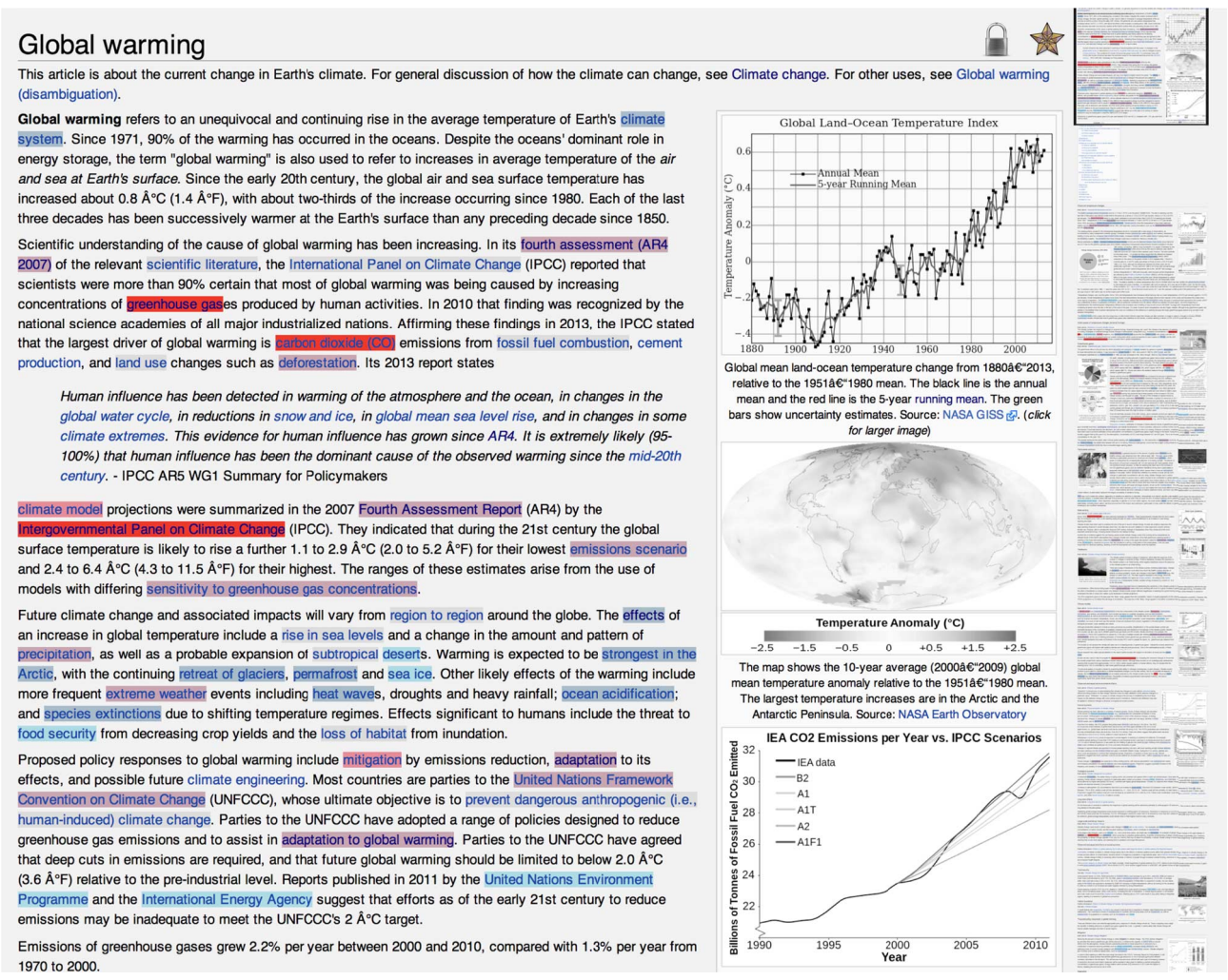

**Figure 2: Controversial topics in the 'global warming' article.**

When clicking on the highlighted controversial wiki links within an article, a list of edits involving, or with debate about, that wiki link unfolds (see Figure 1). This edit table adopts a classic diff visualization that shows the changed and deleted parts with color-coding. It also shows several other relevant variables like the id of the revision in which the edit was made, the name of the editor, the edit summary, the section in which the edit was made, and the timestamp of the edit.

Since the displayed edits only show the sentences in which the wiki link resides, in contrast with the display of all edits in a revision – as offered by the original Wikipedia interface, this function allows the user to zoom in on a particular wiki link to analyze its controversial history. The edit table thus allows the user to further scrutinize the construction of controversial topics in an article and to gain an understanding of what precisely has been disputed about that wiki link as well as which positions were taken.

### Ranking Disputed Topics

The second view, which we call the *controversy dashboard* (see Figure 3), is designed to provide a more analytical view on the data. The dashboard consists of a table listing all the controversial elements, ranked from the most controversial to the least one. Note that the list can also include elements that were removed from the article (e.g. in the article on 'Abortion' the highly controversial wiki link 'death' was removed in 2011). Such elements are struck through (see Figure 3).

Similar to the layer view, the bar on the left represents how controversial a wiki link is through its associated color. A timeline shows the amount of all edits through time, allowing the user to identify historical periods where several edits affected the same element. When the timeline is clicked, the edit table shows the edits associated to that wiki link (see Figure 1). Underneath the timeline one may find a colored bar indicating at what point in time most controversial edits were made. The higher the share of substantive disagreeing edits made to the wiki link in a particular month, the redder the color for that month. This allows one to quickly locate when the link was disputed most. The same colors are used as in the layer view and the overall ranking of the wiki links; grey is added to indicate the absence of controversial edits.

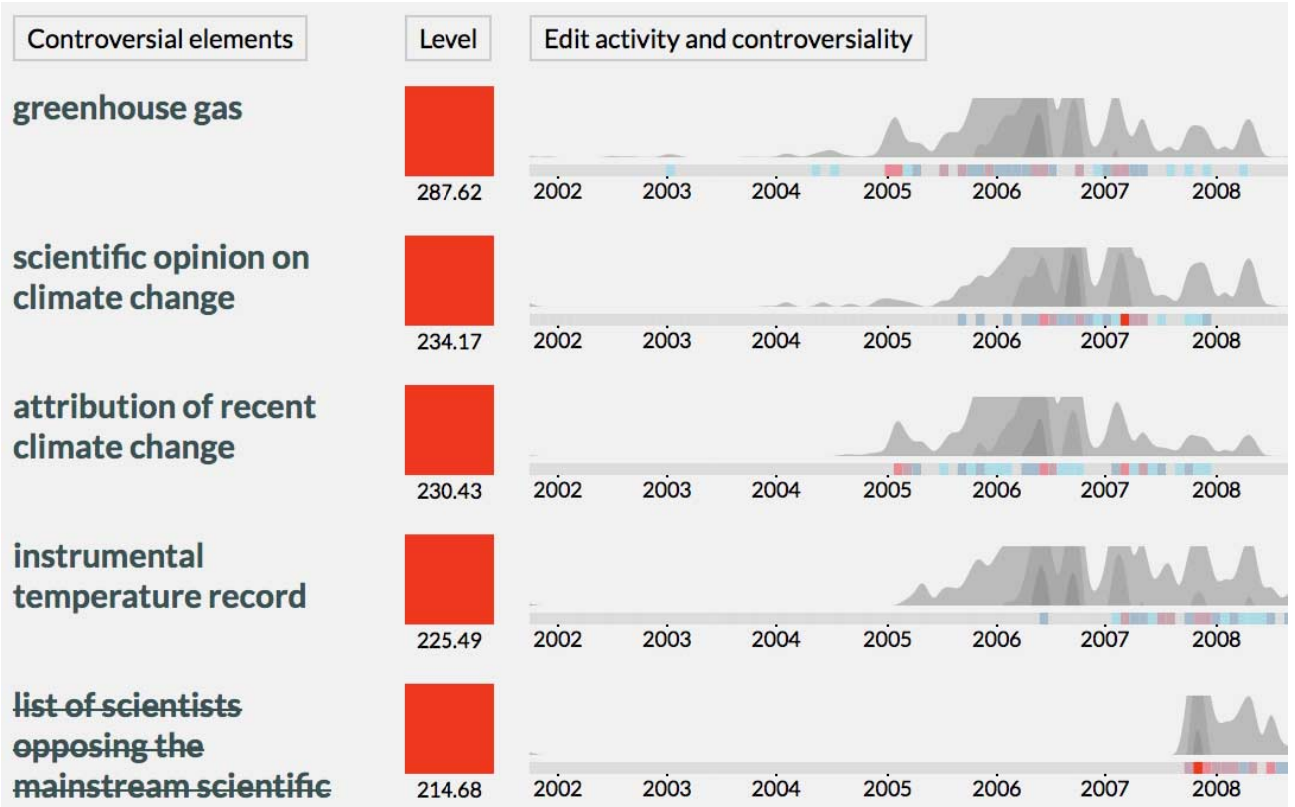

**Figure 3: Detail of dashboard view of the most controversial wiki links in the 'global warming' article on 1 August 2014.**

Both views are accessible via a menu where one can choose an article and the time frame of the analysis. One can thus explore what was more controversial in a given article in, say, 2007. Or what has become controversial since then. It allows one to further study the trajectory of which wiki links were controversial at which time, i.e. at what point a topic was still 'hot' or whether its controversialness has 'cooled down' over time.

When an article is created and news is still uncertain, there may be disputes about what is happening and articles may contain many controversial topics early in their life cycle. That does not mean that those disputes cannot get resolved or that others cannot emerge, as the article on the 'Fukushima Daiichi nuclear disaster' exemplifies. Whereas at its start the main disputes revolved around the accuracy of radioactive measurements and the comparison with the Chernobyl disaster, two years later the disputes concern health effects. By aggregating edit discussions around wiki links our approach thus allows both timely and core controversies to surface.

## RELATED WORK

Conflicts in Wikipedia have previously been studied by observing article edit histories [5], by considering reverts [4, 13], by analyzing talk pages [7, 9], or a combination of the aforementioned [14]. Most of these studies were primarily focused on the social dynamics between editors and only few quantitative approaches have taken into account what the controversy is about, as we do.

Related controversy research has found Wikipedia article sections as the main source of dispute [11]. Our method has a finer granularity and focuses on wiki links as the loci of dispute. Others have similarly regarded wiki links as indications of the subject composition of an article [1, 3, 6].

Additional research has characterized and visualized conflict and coordination on Wikipedia [10, 8] and identified which articles are controversial [16, 17]. While the latter uses articles as the basic units of content to identify which articles are controversial, to our knowledge no research has been pursued in identifying which specific topics within an article are most controversial. Although Viégas et al. [15] and Adler et al. [2] show which content remains stable within an article, we are interested in the opposite: which content has been contested the most?

## CONCLUSION AND DISCUSSION

Recognizing the potential of Wikipedia's edit histories for providing insight into societal controversies, and recognizing that each link within an article can be seen as a focal point for debate, has allowed us to use Wikipedia articles as interesting sites to map controversies. Wiki links identify the main topics in an article and in this paper we have shown how they may be assigned controversy scores by associating wiki links to substantive, disagreeing, edits and then weighting and counting them. We also introduced a tool with two views that are intended to get a quick overview of what is, or was, controversial in an article, why, where, when and to what extent.

The design, and accompanying open-source demo[5], furthermore allow the users of our tool, Contropedia, to zoom into the specific changes around wiki links in the unfolding debate. During case studies, various social researchers found it an invaluable tool to visualize the dynamics of techno-scientific and other societal debates as they unfold on Wikipedia, displaying the framing and phrasing of issues, and helping to clarify conflicts about the content of an article[6]. We think this software may prove useful for Wikipedians as well, as it allows them to gain insight into the substance and build-up of controversies, and allows them to make informed decisions when managing edit wars and disagreements about the articles' content.

We are further developing the tool presented here as part of an elaborate toolkit to study social life on Wikipedia. We are currently associating discussion threads with wiki links to provide further detail about conflict surrounding wiki links. We are also experimenting with calculating the controversy scores using only the parts of sentences involved in substantial edits (instead of using entire sentences). Furthermore, we intend to use other focal points for measuring dispute such as external links, references, figures, and templates. Using these types of elements our algorithm remains language agnostic and has substantial computational savings compared to text-based approaches.

## ACKNOWLEDGEMENTS

This research was supported by EU FP7 EINS grant #288021 and in part by EU FP7 EMAPS grant #288964.

## REFERENCES

1. S. Adafre & M. de Rijke. Discovering missing links in Wikipedia. In *Proc. LinkDD*, 2005.
2. B. Adler et al. Assigning trust to Wikipedia content. In *Proc. WikiSym*, 2008.
3. Bao, P. et al. Omnipedia: Bridging the Wikipedia Language Gap. In *Proc. CHI,* 2012.
4. U. Brandes & J. Lerner. Visual analysis of controversy in user-generated encyclopedias. *Information Visualization*, 7(1):34–48. 2008.
5. M. Ekstrand & J. Riedl. rv you're dumb: identifying discarded work in wiki article history. In *Proc. WikiSym*, 2009.
6. B. Hecht & D. Gergle. Measuring self-focus bias in community-maintained knowledge repositories. In *Proc. C&T*, 2009.
7. A. Kaltenbrunner & D. Laniado. There is No Deadline-Time Evolution of Wikipedia Discussions. In *Proc. WikiSym*, 2012.
8. A. Kittur, B. Suh, B. Pendleton & E. Chi. He says, she says: conflict and coordination in Wikipedia. In *Proc. SIGCHI*, 2007.
9. D. Laniado et al. When the Wikipedians talk: network and tree structure of Wikipedia discussion pages. In *Proc. ICWSM*, 2011.
10. C. Li, A. Datta & A. Sun. Mining latent relations in peer-production environments: a case study with Wikipedia article similarity and controversy. *Social Network Analysis and Mining*, 2(3):265–278, 2012.
11. H. Rad & D. Barbosa. Identifying controversial articles in Wikipedia: A comparative study. In *Proc. Wikisym*, 2012.
12. J. Reagle. The Argument Engine, In *Critical point of view: A Wikipedia Reader*. G. Lovink and N. Tkacz (eds). INC Amsterdam, 2011.
13. B. Suh et al. Us vs. them: Understanding social dynamics in Wikipedia with revert graph visualizations. In *Proc. VAST*, 2007.
14. R. Sumi et al. Edit wars in Wikipedia. In *Proc. PASSAT*, 2011.
15. F. Viégas, M. Wattenberg & K. Dave. Studying cooperation and conflict between authors with history flow visualizations. In *Proc. SIGCHI*, 2004.
16. B. Vuong et al. On ranking controversies in Wikipedia: models and evaluation. In *Proc. WSDM*, 2008.
17. T. Yasseri et al. Dynamics of conflicts in Wikipedia. *PloS one*, 7(6):e38869, 2012.

[5] http://www.contropedia.net

[6] See http://contropedia.net/#case-studies